\documentclass[letterpaper, 10 pt, conference]{ieeeconf}  

\IEEEoverridecommandlockouts                              
                                                          
\usepackage{amsthm}
\usepackage{cite}
\usepackage{graphicx}
\usepackage{textcomp}
\usepackage{xcolor}
\usepackage{comment}
\usepackage{amssymb}
\usepackage{siunitx}
\usepackage{soul}
\usepackage{subcaption}
\usepackage{siunitx}
\usepackage[font=footnotesize]{caption}
\usepackage{hyperref} 
\usepackage{mathtools}
\usepackage[ruled,vlined,linesnumbered]{algorithm2e} 

\SetCommentSty{mycommfont}
\usepackage{bm}
\usepackage{mdwmath}
\usepackage{times} 
\usepackage{blkarray}
\usepackage{amsmath}
\usepackage{booktabs}  
\usepackage{makecell}
\usepackage{nomencl}
\usepackage{float}

\newtheorem{defn}{Definition}[section]

\newcommand{\Zeta}{\mathrm{Z}}

\begin{document}
\bstctlcite{IEEEexample:BSTcontrol} 

\title{\LARGE 
Walking-by-Logic: Signal Temporal Logic-Guided Model Predictive Control for Bipedal Locomotion Resilient to External Perturbations}
\author{Zhaoyuan Gu, Rongming Guo, William Yates, Yipu Chen, and Ye Zhao
\thanks{The authors are with the Laboratory for Intelligent Decision and Autonomous Robots, Woodruff School of Mechanical Engineering, Georgia Institute of Technology. {\tt\footnotesize \{zgu78, rguo61, wyates7, ychen3302, yezhao\}@gatech.edu}}
}

\maketitle
\begin{abstract}
This study proposes a novel planning framework based on a model predictive control formulation that incorporates signal temporal logic (STL) specifications for task completion guarantees and robustness quantification. This marks the first-ever study to apply STL-guided trajectory optimization for bipedal locomotion push recovery, where the robot experiences unexpected disturbances. Existing recovery strategies often struggle with complex task logic reasoning and locomotion robustness evaluation, making them susceptible to failures caused by inappropriate recovery strategies or insufficient robustness.
To address this issue, the STL-guided framework generates optimal and safe recovery trajectories that simultaneously satisfy the task specification and maximize the locomotion robustness. 
Our framework outperforms a state-of-the-art locomotion controller in a high-fidelity dynamic simulation, especially in scenarios involving crossed-leg maneuvers. Furthermore, it demonstrates versatility in tasks such as locomotion on stepping stones, where the robot must select from a set of disjointed footholds to maneuver successfully.

\end{abstract}

\section{Introduction}
This study investigates signal temporal logic (STL) based formal methods for robust bipedal locomotion, with a specific focus on circumstances where a robot encounters environmental perturbations at unforeseen times.

Robust bipedal locomotion has been a long-standing challenge in the field of robotics. 
While existing works have achieved impressive performance using reactive regulation of angular momentum \cite{MIT_ICRA22, MomentumController} or predictive control of foot placement \cite{MIT_CBF, RMP}, few offer formal guarantees on a robot's ability to recover from perturbations, a feature considered crucial for the safe deployment of bipedal robots. To this end, our research centers around designing task specifications for bipedal locomotion push recovery, and employing trajectory optimization that assures task correctness and guarantees system robustness.

Formal methods for bipedal systems have gained significant attention in recent years \cite{LTL_Nav_Kulgod, LTL_Nav_Warnke}. 
The prevailing approach in existing works often relies on abstraction-based methods such as linear temporal logic (LTL) \cite{HadasSynthesis2018} with relatively simple verification processes, which abstract complex continuous behaviors into discrete events and low-dimensional states. 
However, challenges arise when addressing continuous, high-dimensional systems like bipedal robots. 
As a distinguished formal logic, STL \cite{STL_MPC_Raman} offers mathematical guarantees of specifications on dense-time, real-valued signals, making it suitable for reasoning about task logic correctness and quantifying robustness in complex robotic systems.



\begin{figure}[t]
\centerline{\includegraphics[width=0.48\textwidth]{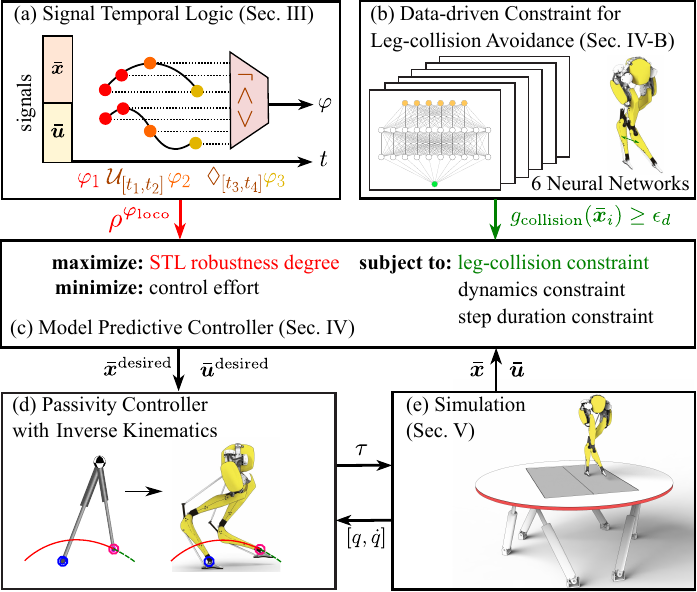}}
\caption{Block diagram of the proposed framework. (a) The signal temporal logic specification $\varphi^{\rm loco}$ specifies the locomotion task. (b) A set of data-driven kinematic constraints enforce the leg self-collision avoidance. (c) The model predictive control-based trajectory optimization solves a stable locomotion trajectory. (d) A whole-body controller tracks the desired trajectory. (e) Perturbed walking experiments on our bipedal robot Cassie.}
\label{fig:framework}
\vspace{-0.15in}
\end{figure}

Self-collision avoidance is another crucial component for ensuring restabilization from disturbances, especially for scenarios involving crossed-leg maneuvers \cite{RMP, MIT_CBF, IHMC_cross} where the distance between the robot's legs diminishes, as shown in Fig.~\ref{fig:framework}(b).
Several previous studies \cite{MomentumController, IHMC_step} relied on inverted pendulum models to plan foot placements for recovery but often overlooked the risk of potential self-collisions during the execution of the foot placement plan.  On the other hand, swing-leg trajectory planning that considers full-body kinematics and collision checking is prohibitively expensive for online computation.


In order to address these challenges, we design an optimization-based planning framework, illustrated in Fig.~\ref{fig:framework}. 
As a core component of the framework, a model predictive controller (MPC) encodes a series of STL specifications (e.g., stability and foot placements) as an objective function to enhance task satisfaction and locomotion robustness. Furthermore, this MPC ensures safety against leg self-collision via a set of data-driven kinematic constraints. 

Solving the MPC generates a reduced-order optimal plan that describes the center of mass (CoM) and swing-foot trajectories, including the walking-step durations. From this MPC trajectory, a low-level controller derives a full-body motion through inverse kinematics and then uses a passivity-based technique for motion tracking. 
We summarize our core contributions as follows:
\begin{itemize}
    \item This work represents the first-ever step towards incorporating STL-based formal methods into trajectory optimization (TO) for dynamic legged locomotion. We design a series of STL task specifications that guide the planning of bipedal locomotion under perturbations. 
    \item We propose a Riemannian robustness metric that evaluates the walking trajectory robustness based on reduced-order locomotion dynamics. The Riemannian robustness is seamlessly encoded as an STL specification and is therefore optimized in the TO for robust locomotion.
    \item We conduct extensive push recovery experiments with perturbations of varying magnitudes, directions, and timings. We compare the robustness of our framework with that of a foot placement controller baseline \cite{MomentumController}.
\end{itemize}

This work is distinct from our previous study \cite{Gu_push} in the following aspects.
(i) Instead of a hierarchical task and motion planning (TAMP) framework using abstraction-based LTL \cite{Gu_push}, this study employs an optimization-based MPC that integrates STL specifications to allow real-valued signals. This property eliminates the mismatch between high-level discrete action sequences and low-level continuous motion plans. 
(ii) 
The degree to which STL specifications are satisfied is quantifiable, enabling the MPC to provide a least-violating solution when the STL specification cannot be strictly satisfied. The LTL-based planner in  \cite{Gu_push}, on the other hand, makes decisions only inside the robustness region, which is more vulnerable in real-system implementation.

\section{Non-periodic Locomotion Modeling}
\label{Sec:LIP}
\subsection{Hybrid Reduced-Order Model for Bipedal Walking}
\label{sec:dynamics}

We propose a new reduced-order model (ROM) that extends the traditional linear inverted pendulum model (LIPM) \cite{Kajita2001, Kajita2003}. The LIPM features a point mass denoted as the center-of-mass (CoM), and a massless telescopic leg that maintains the CoM at a constant height. The LIPM  has a system state $\boldsymbol{x} \coloneqq [\boldsymbol{p}_{\rm CoM}; \boldsymbol{v}_{\rm CoM}]$, where $\boldsymbol{p}_{\rm CoM} = [p_{{\rm CoM}, x}; \;p_{{\rm CoM}, y}; \;p_{{\rm CoM}, z}]$ and $ \boldsymbol{v}_{\rm CoM} = [v_{{\rm CoM}, x}; \;v_{{\rm CoM}, y}; \;v_{{\rm CoM}, z}]$ are the position and velocity of the CoM in the local stance-foot frame, as shown in Fig.~\ref{fig:specification}(a). The LIPM dynamics are expressed as follows: \begin{equation}\label{eq:lipm}
\left[ \begin{matrix}
      {\ddot{p}}_{{\rm CoM},x}\\
      {\ddot{p}}_{{\rm CoM},y}
\end{matrix} \right]
= 
\omega^2 
\left[ \begin{matrix}
      {p}_{{\rm CoM},x}\\
      {p}_{{\rm CoM},y}
\end{matrix} \right]
\end{equation}
where $\omega = \sqrt{\textsl{g}/{p}_{{\rm CoM},z}}$ and $\textsl{g}$ is the acceleration due to gravity. The subscripts $x$ and $y$ indicate the sagittal and lateral components of a vector, respectively.

We design a variant of the traditional LIPM that additionally models the swing-foot position and velocity (Fig.~\ref{fig:specification}(a)). In effect, the state vector is augmented as $\boldsymbol{\bar{x}} \coloneqq [\boldsymbol{p}_{\rm CoM}; \boldsymbol{v}_{\rm CoM}; \boldsymbol{p}_{\rm swing}], \boldsymbol{p}_{\rm swing} \in \mathbb{R}^{3}$, and the control input $\boldsymbol{\bar{u}}$ sets the swing foot velocity $\boldsymbol{\dot{p}}_{\rm swing}$.
Moreover, we define $\boldsymbol{y} = [\boldsymbol{\bar{x}}; \boldsymbol{\bar{u}}] \in \mathbb{R}^{12}$ as the system output, which will be used in Sec.~\ref{Sec:STL} for signal temporal logic (STL) definitions. 

At contact time, a \textit{reset map} $\boldsymbol{\bar{x}}^+ = \bar{\Delta}_{j \rightarrow j+1} (\boldsymbol{\bar{x}}^{-}) $ uses the swing foot location to transition to the next walking step: 
\begin{equation}
\left[ \begin{matrix}
      \boldsymbol{p}_{\rm CoM}^+\\
      \boldsymbol{v}_{\rm CoM}^+\\
      \boldsymbol{p}_{\rm swing}^+
\end{matrix} \right] 
= 
\left[ \begin{matrix}
      \boldsymbol{p}_{\rm CoM}^- - \boldsymbol{p}_{\rm swing}^-\\
      \boldsymbol{v}_{\rm CoM}^- \\
      - \boldsymbol{p}_{\rm swing}^-
\end{matrix} \right]
\end{equation}
This occurs when the system state reaches the switching condition $\mathcal{S} \coloneqq \{\bar{\boldsymbol{x}}| {p}_{{\rm swing},z} = h_{\rm terrain} \}$, where $h_{\rm terrain}$ is the terrain height. Note that the aforementioned position and velocity parameters are expressed in a local coordinate frame attached to the stance foot. The swing foot becomes the stance foot immediately after it touches the ground. \\

\vspace{-0.1in}
\noindent\textbf{Remark.} 
\textit{Our addition of the swing-foot position  $\boldsymbol{p}_{{\rm swing}}$, together with $\boldsymbol{p}_{\rm CoM}$, uniquely determines the leg configuration of the Cassie robot (e.g., via inverse kinematics), allowing us to plan a collision-free trajectory using only the ROM in Sec.~\ref{sec:MLP}. }

\subsection{Keyframe-Based Non-Periodic Locomotion and Riemannian Robustness} 
\label{sec:keyframe_loco}

To enable robust locomotion that adapts to unexpected perturbations or rough terrains, we employ the concept of a \textit{keyframe} (proposed in our previous work \cite{Zhao2017IJRR}) as a critical locomotion state. The keyframe summarizes a non-periodic walking step in a reduced-order space, and it addresses the robot's complex interaction with the environment. The keyframe allows for the quantification of locomotion robustness, which will be integrated as a cost function within the trajectory optimization in Sec.~\ref{Sec:MPC}.

\begin{figure}[t]
\centering
\includegraphics[width=0.45\textwidth]{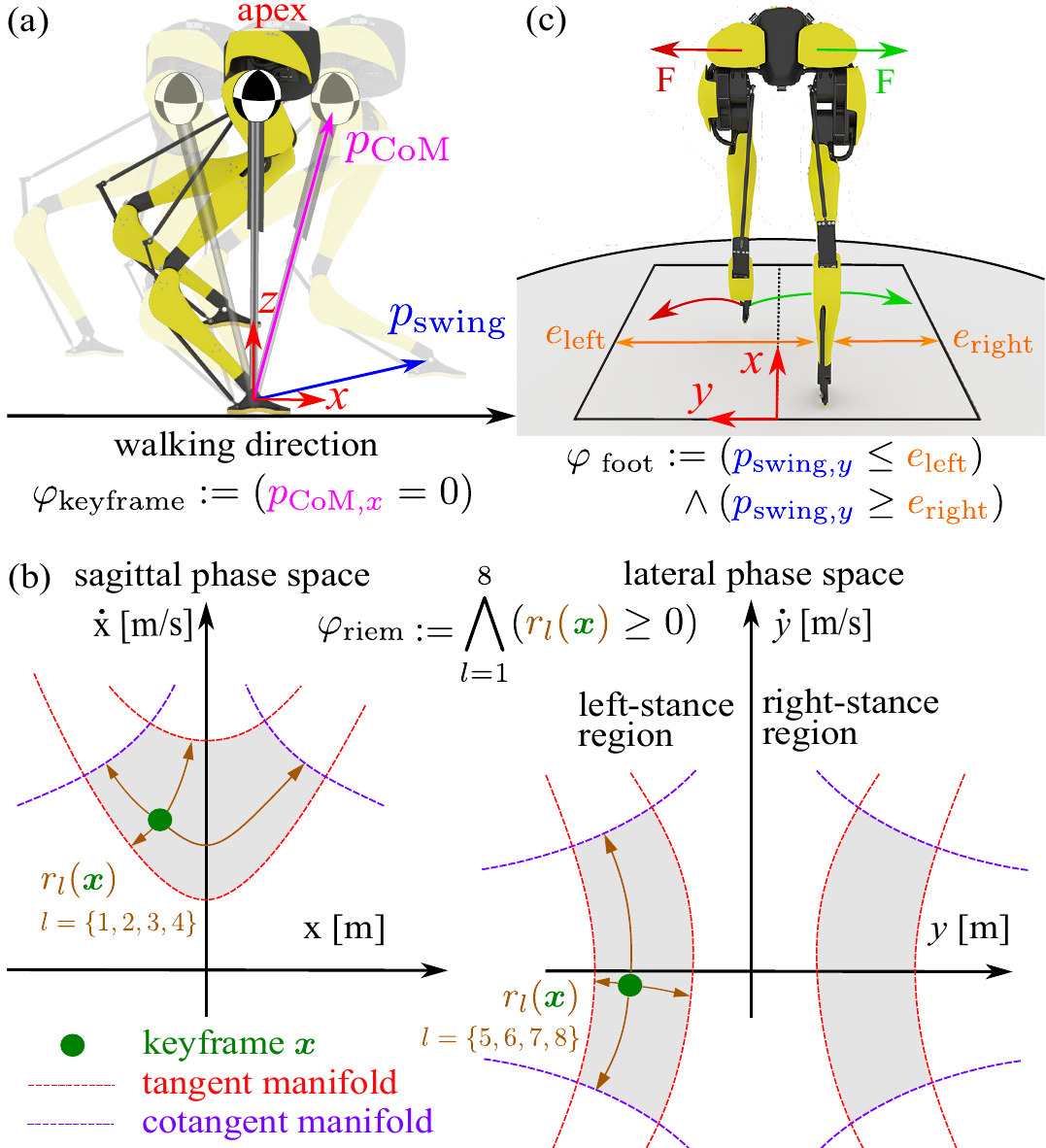}
\caption{Illustration of the locomotion specifications. (a) The highlighted state in the middle is the keyframe of a walking step. (b) The grey areas are the Riemannian regions in the sagittal and lateral phase spaces. The signed distances to the bounds of the Riemannian regions are indicated by the arrows. (c) Cassie's foot is specified to step inside the lateral bounds.}
\label{fig:specification}
\vspace{-0.2in}
\end{figure}

\begin{defn}[Locomotion keyframe]\label{def:keyframe}
Locomotion keyframe is defined as
the robot's CoM state $(\boldsymbol{p}_{\rm CoM}, \boldsymbol{v}_{\rm CoM})$ at the apex, i.e., when the CoM is over the stance foot in the sagittal direction 
(${p}_{{\rm CoM}, x} = 0$), as shown in Fig.~\ref{fig:specification}(a).
\end{defn}

To quantify the robustness of a non-periodic walking step, we design a robust region centered around a nominal keyframe state in a Riemannian space. The Riemannian space \cite{Zhao2017IJRR} is a reparameterization of the Euclidean CoM phase space using tangent and cotangent locomotion manifolds, represented by a pair $({\sigma}, {\zeta})$. ${\sigma}$ represents the tangent manifold along which the CoM dynamics evolve, while ${\zeta}$ represents the cotangent manifold orthogonal to ${\sigma}$. These manifolds can be derived analytically from the LIPM dynamics in (\ref{eq:lipm}); the detailed derivation is in \cite{Zhao2017IJRR}. 
Within the Riemannian space, we define a robust keyframe region that enables stable walking. This region is referred to as the Riemannian region.

\begin{defn}[Riemannian region]\label{def:riem}
The Riemannian region $\mathcal{R}$ is the area centered around a nominal keyframe state $(\sigma_{\rm nom}, \zeta_{\rm nom})$: $\mathcal{R}_{d} \coloneqq \{ (p_{{\rm CoM},d}, v_{{\rm CoM},d}) \; |\; \allowbreak \sigma(p_{{\rm CoM},d}, v_{{\rm CoM},d}) \in \Sigma_d, \; \zeta(p_{{\rm CoM}, d}, v_{{\rm CoM}, d})\in \Zeta_d
 \}$,
\noindent where $d \in \{x, y\}$ indicates sagittal and lateral directions, respectively. 
$\Sigma_d = [\sigma_{{\rm nom},d}-\delta \sigma_d, \sigma_{{\rm nom},d}+\delta \sigma_d]$ and $\Zeta_d = [\zeta_{{\rm nom},d}-\delta \zeta_d, \zeta_{{\rm nom},d}+\delta \zeta_d]$ are the ranges of the manifold values for $\sigma$ and $\zeta$, where $\delta \sigma_d, \delta \zeta_d$ are robustness margins. 
\end{defn}

The sagittal and lateral Riemannian regions in the phase space are illustrated in Fig.~\ref{fig:specification}(b) as shaded areas. The bounds of these Riemannian regions are curved in the phase space because they obey the LIPM locomotion dynamics. 
Notably, while two Riemannian regions exist in the lateral phase space, only one is active at any given time, corresponding with the stance leg labeled in Fig.~\ref{fig:specification}(b).

\begin{defn}[Riemannian robustness]\label{def:riem_robust}
The Riemannian robustness $\rho_{\rm riem}$ is the minimum signed distance of an actual keyframe CoM state $\boldsymbol{x}$ to all the bounds of the Riemannian regions. Namely,
$\rho_{\rm riem} := {\rm min}_{l=1}^8(r_l(\boldsymbol{x}))$, where $r_l(\boldsymbol{x})$ is the signed distance to the $l^{\rm th}$ bound of the Riemannian regions, as illustrated in Fig.~\ref{fig:specification}(b). We have a total of $8$ bounds as the sagittal and lateral Riemannian regions each have $4$ bounds. 
\end{defn}
Riemannian robustness represents the locomotion robustness in the form of Riemannian regions. Any keyframe inside the Riemannian region has a positive robustness value, which indicates a stable walking step. 
In the next section, our goal is to leverage Riemannian robustness as an objective function and use STL-based optimization to plan robust trajectories for locomotion recovery.


\section{Signal Temporal Logic and Task Specification for Locomotion}
\label{Sec:STL}
Signal temporal logic (STL) \cite{STL_Origin} uses logical symbols of negation ($\neg$), conjunction ($\wedge$), and disjunction ($\vee$), as well as temporal operators such as eventually ($\Diamond$), always ($\square$), and until ($\mathcal{U}$) to construct specifications. A specification is defined with the following syntax:
\begin{equation}\begin{split}
\varphi \coloneqq \;
& \pi \;|\; \neg\varphi \;|\; \varphi_1 \wedge \varphi_2 \;|\; \varphi_1 \vee \varphi_2 \;|\; \\ 
& \Diamond_{[t_1,t_2]}\;\varphi \;|\; \square_{[t_1,t_2]}\;\varphi \;|\; 
\varphi_1 \; \mathcal{U}_{[t_1,t_2]}\; \varphi_2
\end{split}\end{equation}
where $\varphi$, $\varphi_1$, and $\varphi_2$ are STL specifications. $\pi := (\mu^\pi(\boldsymbol{y}) - c \geq 0)$ is a boolean predicate, where $\mu^\pi: \mathbb{R}^p \to \mathbb{R}$ is a vector-valued function, $c \in \mathbb{R}$, and the signal $\boldsymbol{y}(t) : \mathbb{R}_+ \to \mathbb{R}^p$ is a $p$-dimensional vector at time $t$. For a dynamical system, the signal $\boldsymbol{y}(t)$ is the system output (in our study, $\boldsymbol{y} = [\boldsymbol{\bar{x}}; \boldsymbol{\bar{u}}] \in \mathbb{R}^{12}$). 
The time bounds of an STL formula are denoted with $t_1$ and $t_2$, where $0 \leq t_1 \leq t_2 \leq t_{\rm end}$ and $t_{\rm end}$ is the end of a planning horizon. The validity of an STL specification is inductively defined using the rules in Table~I.

\begin{table}[H]
\centering
TABLE I \\
VALIDITY SEMANTICS OF SIGNAL TEMPORAL LOGIC
\begin{tabular}{l c c} \\ 
\hline

$(\boldsymbol{y},t) \models \pi$ & $\Leftrightarrow$ & $\mu ^ \pi  (\boldsymbol{y}(t)) - c \geq 0$ \\

$(\boldsymbol{y},t) \models \neg\varphi$ & $\Leftrightarrow$ & $(\boldsymbol{y},t) \not\models \varphi$ \\

$(\boldsymbol{y},t) \models \varphi_1 \wedge \varphi_2$ & $\Leftrightarrow$ & $(\boldsymbol{y},t) \models \varphi_1 \wedge (\boldsymbol{y},t) \models \varphi_2$ \\

$(\boldsymbol{y},t) \models \varphi_1 \vee \varphi_2$ & $\Leftrightarrow$ & $(\boldsymbol{y},t) \models \varphi_1 \vee (\boldsymbol{y},t) \models \varphi_2$ \\

$(\boldsymbol{y},t) \models \Diamond_{[t_1,t_2]}\varphi$ & $\Leftrightarrow$ & $\exists {t^{'}\in[t+t_1,t+t_2]}, (\boldsymbol{y},t^{'}) \models \varphi$ \\

$(\boldsymbol{y},t) \models \square_{[t_1,t_2]}\varphi$ & $\Leftrightarrow$ & $\forall {t^{'}\in[t+t_1,t+t_2]}, (\boldsymbol{y},t^{'}) \models \varphi$\\

$(\boldsymbol{y},t) \models {\varphi_1}\mathcal{U}_{[t_1,t_2]}{\varphi_2}$ & $\Leftrightarrow$ & \makecell{$\exists {t^{'}\in[t+t_1,t+t_2]}, (\boldsymbol{y},t^{'}) \models \varphi_2 \wedge $\\$ \forall {t^{''}\in[t+t_1,t^{'}]} (\boldsymbol{y},t^{''}) \models \varphi_1$} \\ 
 
\hline
\end{tabular}
\end{table}
\label{tab:STL_satisfy}

STL provides the capability of quantifying \textit{robustness degree} \cite{MTL_Papas} \cite{Belta_STL_review}. A positive robustness degree indicates specification satisfaction, and its magnitude represents the resilience to disturbances without violating this specification. 
When incorporated into trajectory optimization as a cost, the robustness degree allows for a minimally specification-violating trajectory if the task specification cannot be satisfied strictly \cite{Robust_STL_MPC}. 
Table~II shows the semantics of the robustness degree. 
\begin{table}[H]
\centering
TABLE II \\
ROBUSTNESS DEGREE SEMANTICS
\begin{tabular}{l c c} \\ 
\hline

$\rho ^ \pi (\boldsymbol{y},t)$ & $=$ & $\mu ^ \pi  (\boldsymbol{y}(t)) - c$ \\

$\rho ^ {\neg\varphi} (\boldsymbol{y},t)$ & $=$ & $-\rho ^ \varphi  (\boldsymbol{y},t)$ \\

$\rho ^ {\varphi_1 \wedge \varphi_2} (\boldsymbol{y},t)$ & $=$ & ${\rm min} (\rho ^ {\varphi_1} (\boldsymbol{y},t),\rho ^ {\varphi_2} (\boldsymbol{y},t))$ \\

$\rho ^ {\varphi_1 \vee \varphi_2} (\boldsymbol{y},t)$ & $=$ & ${\rm max} (\rho ^ {\varphi_1} (\boldsymbol{y},t),\rho ^ {\varphi_2} (\boldsymbol{y},t))$ \\

$\rho ^ {\Diamond_{[t_1,t_2]}\varphi}(\boldsymbol{y},t)$ & $=$ & ${\rm max}_{t^{'}\in[t+t_1,t+t_2]} (\rho^{\varphi} (\boldsymbol{y},t^{'}))$ \\

$\rho ^ {\square_{[t_1,t_2]}\varphi}(\boldsymbol{y},t)$ & $=$ & ${\rm min}_{t^{'}\in[t+t_1,t+t_2]} (\rho^{\varphi} (\boldsymbol{y},t^{'}))$\\

$\rho ^ {{\varphi_1}\mathcal{U}_{[t_1,t_2]}{\varphi_2}} (\boldsymbol{y},t)$ & $=$ & \makecell{${\rm max}_{t^{'}\in[t+t_1,t+t_2]}({\rm min}(\rho^{\varphi_2}(\boldsymbol{y},t^{'}),$\\${\rm min}_{t^{''}\in[t+t_1,t^{'}]}(\rho^{\varphi_1}(\boldsymbol{y},t^{''}))))$} \\ 
 
\hline
\end{tabular}
\end{table}
\label{tab:robustness}

The rest of this section introduces the locomotion specification $\varphi_{\rm loco}$, designed to guarantee stable walking trajectories. We interpret locomotion stability as a \textit{liveness} property in the sense that a keyframe with a positive Riemannian robustness will \textit{eventually} occur in the planning horizon.

\textit{Keyframe specification $\varphi_{\rm keyframe}$}: 
To enforce properties on a keyframe, we first describe it using an STL formula $\varphi_{\rm keyframe}$, checking whether or not a signal $\boldsymbol{y}$ is a keyframe.
According to Def.~\ref{def:keyframe}, the keyframe occurs when the CoM is over the foot contact in the sagittal direction. Illustrated in Fig.~\ref{fig:specification}(a), this definition is formally specified as $\varphi_{\rm keyframe}:= (\mu^\pi_{{\rm CoM}, x}(\boldsymbol{y}) = 0)$, where the predicate denotes the sagittal CoM position $\mu^\pi_{{\rm CoM}, x}(\boldsymbol{y}) = {p}_{{\rm CoM},x}$. 

\textit{Riemannian robustness $\varphi_{\rm riem}$}: A stable walking step has a keyframe with positive Riemannian robustness; i.e., the keyframe resides in the Riemannian region, as defined in Def.~\ref{def:riem_robust}.
As shown in Fig.~\ref{fig:specification}(b), we encode the Riemannian robustness specification $\varphi_{\rm riem}$ such that it is \textsf{True} when a CoM state $\boldsymbol{x}$ of a signal is inside the Riemannian region:
$\varphi_{\rm riem} := \bigwedge_{l=1}^{8} (r_l(\boldsymbol{x}) \ge 0)$, 
where $r_l(\boldsymbol{x})$ is the signed distance from $\boldsymbol{x}$ to the $l^{\rm th}$ bound of the Riemannian region in the Riemannian space. 

\textit{Locomotion stability $\varphi_{\rm stable}$}:
To encode this property using STL, we specify that the keyframe of the last walking step falls inside the corresponding Riemannian region. This stability property is encoded as $\varphi_{\rm stable} := \Diamond_{[T_{\rm contact}^N,T_{\rm contact}^{N+1}]} (\varphi_{\rm keyframe} \wedge \varphi_{\rm riem})$, 
where $T_{\rm contact}^N$ and $T_{\rm contact}^{N+1}$ are the $N^{\rm th}$ and $N+1^{\rm th}$ contact times and represent the time bounds of the last walking step in the planning horizon. 

\textit{Swing foot bound $\varphi_{\rm foot}$}: For locomotion in a narrow space (e.g., a treadmill, as shown in Fig.~\ref{fig:specification}(c)), we use a \textit{safety} specification $\square{\varphi_{\rm foot}}$ to ensure the foothold lands inside of the treadmill’s edges. The operator $\square$ without a time bound means the specification should hold for the entire planning horizon. We define $\varphi_{\rm foot} := (\mu^\pi_{\rm left}(\boldsymbol{y}) \ge 0) \wedge (\mu^\pi_{\rm right}(\boldsymbol{y}) \ge 0)$, where $\mu^\pi_{\rm left} = -p_{{\rm swing}, y} + e_{\rm left}$ and $\mu^\pi_{\rm right} = p_{{\rm swing}, y} - e_{\rm right}$ are the predicates for limiting the lateral foot location against the left edge $e_{\rm left}$ and the right edge $e_{\rm right}$ of the treadmill.

\textit{Overall locomotion specification $\varphi_{\rm loco}$}: The compounded locomotion specification is $\varphi_{\rm loco} = \varphi_{\rm stable} \wedge (\square \varphi_{\rm foot})$. Satisfying the specification $\varphi_{\rm loco}$ is equivalent to having a positive robustness degree: 
$(\boldsymbol{y},t) \models \varphi_{\rm loco} \Leftrightarrow \rho^{\varphi_{\rm loco}}(\boldsymbol{y},t) \geq 0$. 
In order to maximize the locomotion robustness, we use the robustness degree $\rho ^ {\varphi_{\rm loco}}$ as an objective function in the trajectory optimization in the following section. 


\begin{figure}[t]
    \centering
    \includegraphics[width=0.48\textwidth]{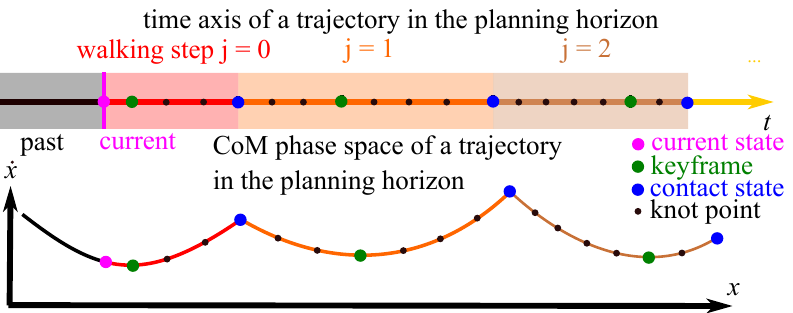}
    \caption{The planning horizon starts from the current measured state (pink). An example of $N = 2$ walking steps and $8$ knot points per walking step is illustrated for simplicity (our actual implementation has $10$ knot points).}
    \label{fig:horizon}
    \vspace{-0.2in}
\end{figure}

\section{Model Predictive Control for Push Recovery}
\label{Sec:MPC}
\subsection{Optimization Formulation}
We design a model predictive controller (MPC) to solve a sequence of optimal states and controls (i.e., signals) that simultaneously satisfy specification $\varphi_{\rm loco}$, system dynamics, and kinematic constraints within an $N$-step horizon. 

The MPC functions as the primary motion planner of the framework and operates in both normal and perturbed locomotion conditions.
Our MPC is formulated as the following nonlinear program:
\begin{align}
\label{eq:MPC}
\min_{\boldsymbol{{X}}, \boldsymbol{{U}}, \boldsymbol{T}} \;\; & w \mathcal{L}(\boldsymbol{{U}}) - \Tilde{\rho}^{\varphi_{\rm loco}}(\boldsymbol{{X}}, \boldsymbol{{U}}) \\
\textrm{s.t.} \quad & \boldsymbol{\bar{x}}_{i+1}^j = f(\boldsymbol{\bar{x}}_{i}^j, \boldsymbol{\bar{u}}_{i}^j, T^j), \qquad i \in \mathbb{H} \setminus \mathbb{S}, & j \in \mathbb{J}  \label{eq:continuous_dynamics}\\
& \boldsymbol{\bar{x}}^{+,j+1} = \bar{\Delta}_{j \rightarrow j+1} (\boldsymbol{\bar{x}}^{-,j}), & j \in \mathbb{J} \label{eq:reset_map}\\
%
%
& g_{\rm collision}(\boldsymbol{\bar{x}}_i) \ge \epsilon_d, & i \in \mathbb{H} \label{eq:collision}\\ 
& g_{\rm duration}(T^j) \ge 0, & j \in \mathbb{J} \label{eq:duration}\\ 
& h_{\rm initial}(\boldsymbol{\bar{x}}_0) = 0, h_{\rm transition}(\boldsymbol{\bar{x}}_i) = 0, & i \in \mathbb{S} \label{eq:initial}
\end{align}
where $\mathbb{H}$ is a set of indices that includes all time steps in the horizon. We design $\mathbb{H}$ to span from the acquisition of the latest measured states till the end of the next $N$ walking steps, with a total of $M$ time steps. Fig.~\ref{fig:horizon} illustrates a horizon with $N=2$. $\mathbb{S}$ is the set of indices containing the time steps of all contact switch events, $\mathbb{S} \subset \mathbb{H}$. $\mathbb{J} = \{0, \ldots, N\}$ is the set of walking step indices. The decision variables include $\boldsymbol{{X}} = \{\boldsymbol{\bar{x}}_1, \ldots, \boldsymbol{\bar{x}}_M\}$, $\boldsymbol{{U}} = \{\boldsymbol{\bar{u}}_1, \ldots, \boldsymbol{\bar{u}}_M \}$, and $\boldsymbol{T} = \{T^0, \ldots, T^N\}$. $\boldsymbol{T}$ is a vector defining the individual step durations for all walking steps. 

$\mathcal{L}(\boldsymbol{{U}}) = \sum_{i=1}^M ||\boldsymbol{\bar{u}}_i||^2$ is a cost function penalizing the control with a weight coefficient $w$. The robustness degree $\Tilde{\rho}^{\varphi_{\rm loco}}(\boldsymbol{{X}},\boldsymbol{{U}})$ represents the degree of satisfaction of the signal $(\boldsymbol{X},\boldsymbol{U})$ with respect to the locomotion specification $\varphi_{\rm loco}$.
$\Tilde{\rho}^{\varphi_{\rm loco}}$ is a smooth approximation of ${\rho}^{\varphi_{\rm loco}}$ using smooth operators \cite{Lin_Smooth}. The exact, non-smooth version ${\rho}^{\varphi_{\rm loco}}$ has discontinuous gradients, which can cause the optimization problem to be ill-conditioned. Maximizing $\Tilde{\rho}^{\varphi_{\rm loco}}(\boldsymbol{{X}},\boldsymbol{{U}})$ encourages the keyframe towards the center of the Riemannian region, as discussed in Sec.~\ref{Sec:STL}. 

To satisfy the LIPM dynamics (\ref{eq:lipm}) while adapting step durations $\boldsymbol{T}$, we use a second-order Taylor expansion to derive the approximated discrete dynamics (\ref{eq:continuous_dynamics}). (\ref{eq:reset_map}) defines the reset map from the foot-ground contact switch. 
(\ref{eq:collision}) represents a set of self-collision avoidance constraints, which ensures a collision-free swing-foot trajectory. The threshold $\epsilon_d$ is the minimum allowable distance for collision avoidance. The $g_{\rm collision}$ is a set of multilayer perceptrons (MLPs) learned from leg configuration data, as detailed in Sec.~\ref{sec:MLP}.  (\ref{eq:duration}) clamps step durations $\boldsymbol{T}$ within a feasible range. By allowing variations in step durations, we enhance the perturbation recovery capability of the bipedal system \cite{step_time_adaptation}. (\ref{eq:initial}) are the equality constraints of the MPC: $h_{\rm initial}$ denotes the initial state constraint; $h_{\rm transition}$ is the guard function posing kinematic constraints between the swing foot height and the terrain height, ${p}_{{\rm swing},z} = h_{\rm terrain}$, for walking step transitions at contact-switching indices in $\mathbb{S}$.



Upon the successful completion of a MPC optimization, the solution is immediately sent to the low-level passivity-based controller \cite{PassivityControl} for tracking and execution. The MPC then reinitializes the same problem based on the latest state measurements. 

\subsection{Data-Driven Self-Collision Avoidance Constraints}
\label{sec:MLP}

\begin{figure}[t]
\centerline{\includegraphics[width=.45\textwidth]{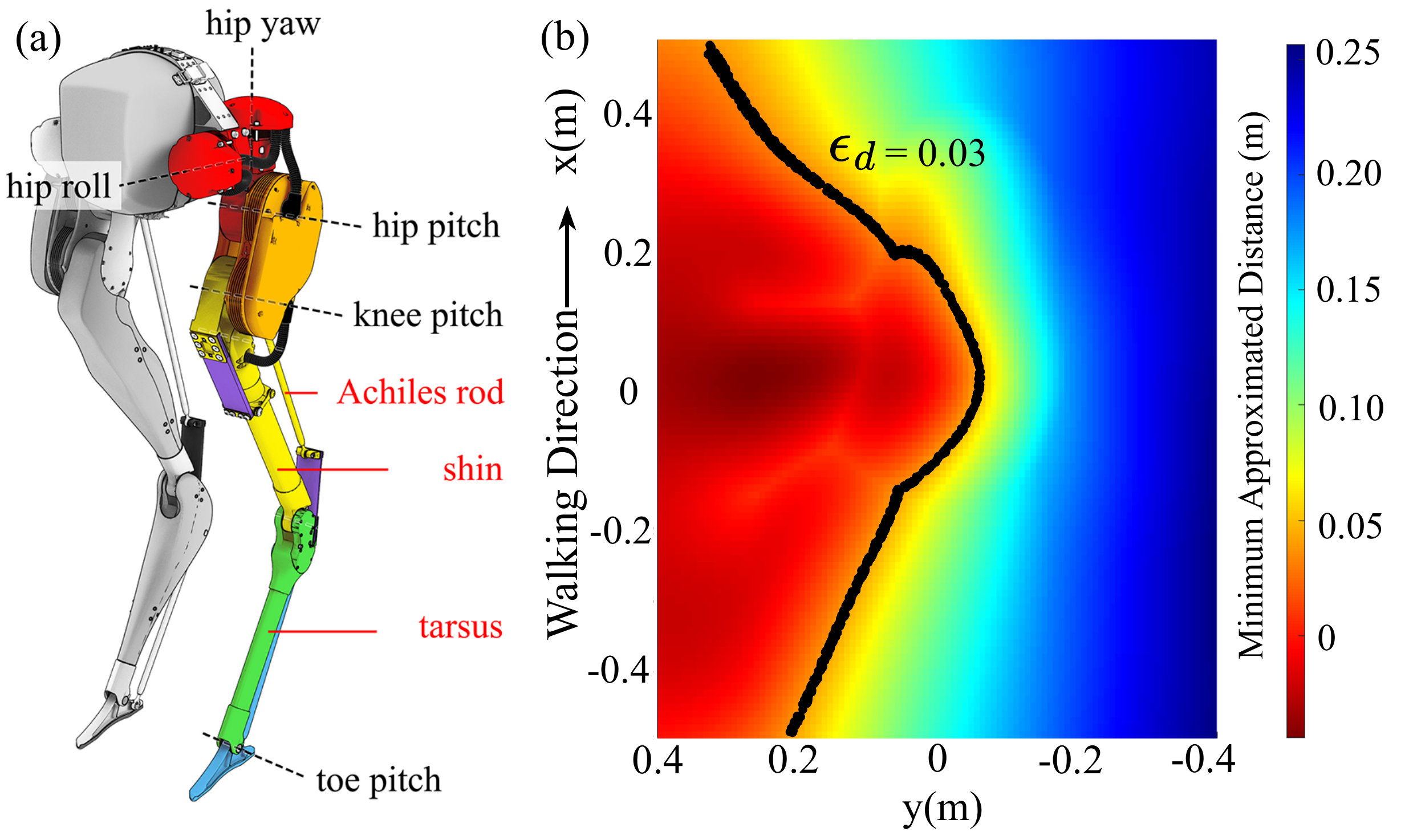}}
\caption{(a) The robot kinematic anatomy for collision pair definitions. (b) The MLP prediction of the minimum distance between Cassie's two legs with the left foot affixed to $(0,0)$ and the right foot moving in the $xy$ plane.}
\label{fig:MLP_Landscape}
\vspace{-0.15in}
\end{figure}

We design a set of MLPs to approximate the mapping from reduced-order linear inverted pendulum model (LIPM) states to the distances between geometry pairs that pose critical collision risks. According to Cassie's kinematic configuration depicted in Fig.~\ref{fig:MLP_Landscape}(a), these pairs include left shin to right shin (LSRS), left shin to right tarsus (LSRT), left shin to right Achilles rod (LSRA), left tarsus to right shin (LTRS), left tarsus to right tarsus (LTRT), and left Achilles rod to right shin (LARS). A total of $6$ MLPs are constructed, each approximating the distance between one geometry pair. The MLPs are then encoded as constraints in the MPC to ensure collision-free trajectories. 

Each MLP consists of $2$ hidden layers of $24$ neurons and is trained on a dataset with $10^6$ entries obtained through an extensive exploration of leg configurations. The MLPs achieved an accurate prediction performance with an average absolute error of $0.002$ m, and an impressive evaluation speed of over $1000$ kHz, compared to $1$ kHz using full-body kinematics-based approaches for collision checking.

We illustrate the effectiveness of the MLPs through kinematic analysis of the collision-free range of motion of Cassie's swing leg during crossed-leg maneuvers. Specifically, we consider a representative crossed-leg scenario where Cassie's left foot is designated as the stance foot and affixed directly beneath its pelvis. We move Cassie's right leg within the $xy$ plane at the same height as the stance foot while recording the minimum value among all $6$ MLP-approximated distances.

The result is plotted as a heat map in Fig. \ref{fig:MLP_Landscape}(b), where the coordinate indicates the location of the swing foot with respect to the pelvis. As expected, the plot reveals a trend of decreasing distance as the swing foot approaches the stance foot. A contour line drawn at $\epsilon_d = 0.03$ m indicates the MLP-enforced boundary between collision-free and collision-prone regions for foot placement. The collision-prone region to the left of the plane exhibits a cluster of red zones, each indicating a different active collision pair. 
\section{Results}
\subsection{Self-Collision Avoidance during Leg Crossing}

We demonstrate the ability of the signal temporal logic-based model predictive controller (STL-MPC) to avoid leg collisions in a critical push recovery setting, where a perturbation forces the robot to execute a crossed-leg maneuver. 

\begin{figure}[t]
\centerline{\includegraphics[width=.45\textwidth]{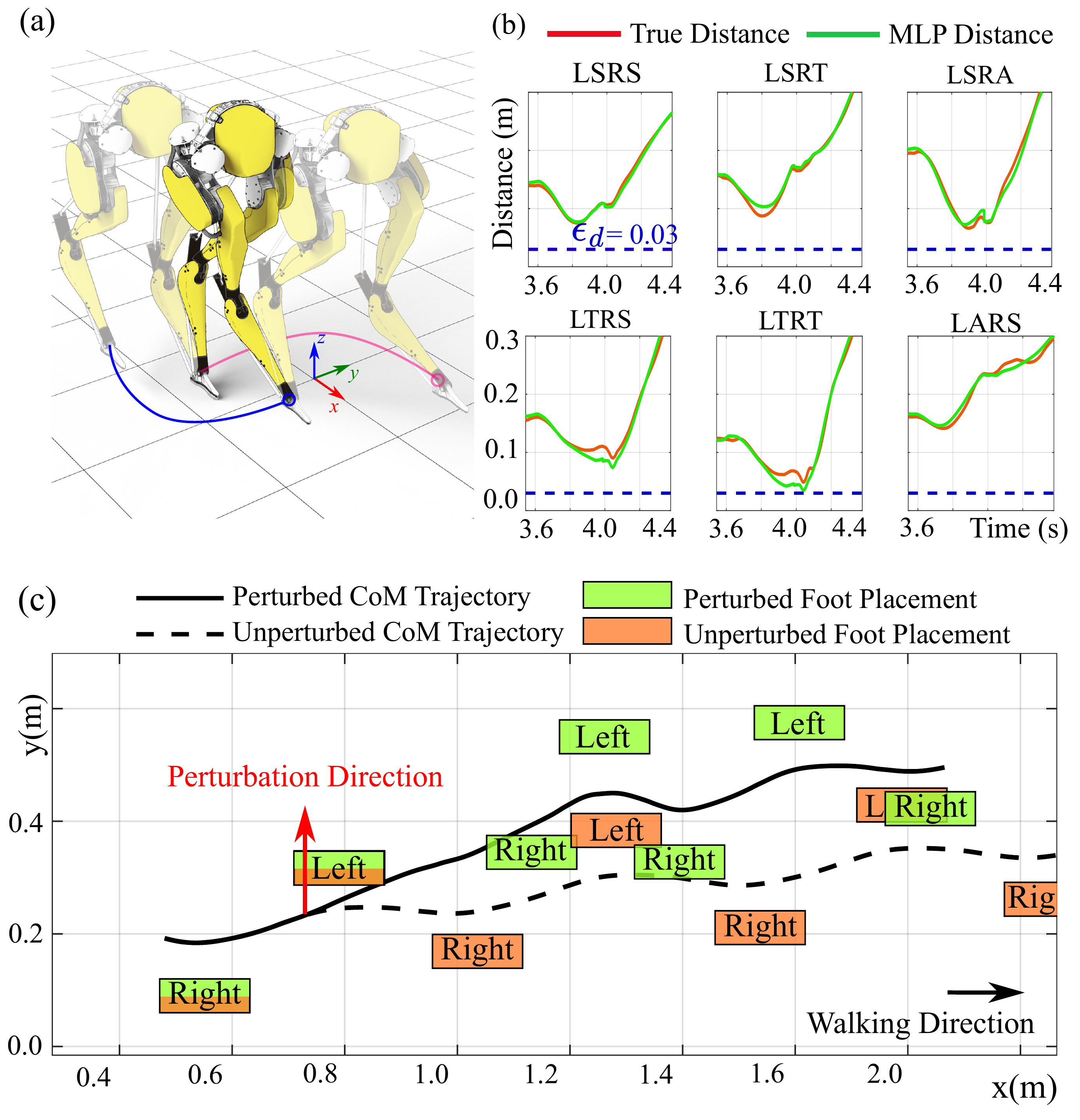}}
\caption{(a) Snapshots of Cassie performing a crossed-leg maneuver for push recovery. (b) The MLP-approximated collision distances are accurate compared with the ground truth, and the planned leg trajectory is safe against the threshold $\epsilon_d = 0.03$. (c) An overhead view of the CoM trajectory and foot placements when a lateral perturbation induces a crossed-leg maneuver.}
\label{fig:collision_avoid}
\vspace{-0.15in}
\end{figure}

The MPC with collision constraints generates a trajectory as shown in Fig.~\ref{fig:collision_avoid}(a), where the swing leg adeptly maneuvers around the stance leg and lands at a safe crossed-leg recovery point. Similarly, the robot extricates itself from the crossed-leg state in the subsequent step, following a curved trajectory that actively avoids self-collisions. An overhead view comparing the perturbed and unperturbed trajectories is shown in Fig.~\ref{fig:collision_avoid}(c). Fig.~\ref{fig:collision_avoid}(b) shows that the multilayer perceptron (MLP)-approximated collision distances are accurate and that the planned trajectory is safe against the threshold $\epsilon_d = 0.03$.

\subsection{Comprehensive, Omnidirectional Perturbation Recovery}

We examine the robustness of the STL-MPC framework through an ensemble of push-recovery tests conducted in simulation, where horizontal impulses are systematically applied to Cassie's pelvis. The impulses are exerted for a fixed duration of $0.1$ s but vary in magnitude, direction, and timing. Specifically, impulses have: $9$ magnitudes evenly distributed between $80$ N and $400$ N; $12$ directions evenly distributed between $0^\circ$ and $330^\circ$; and $4$ locomotion phases at a percentage $s$ through a walking step, where $s = 0\%, \,25\%, \,50\%, \,75\%$. Collectively, this experimental design encompasses a total of 432 distinct trials. For a baseline comparison, the same perturbation procedure is applied to an angular-momentum-based reactive controller (ALIP controller) \cite{MomentumController}.

\begin{figure}[t]
\centerline{\includegraphics[width=.43\textwidth]{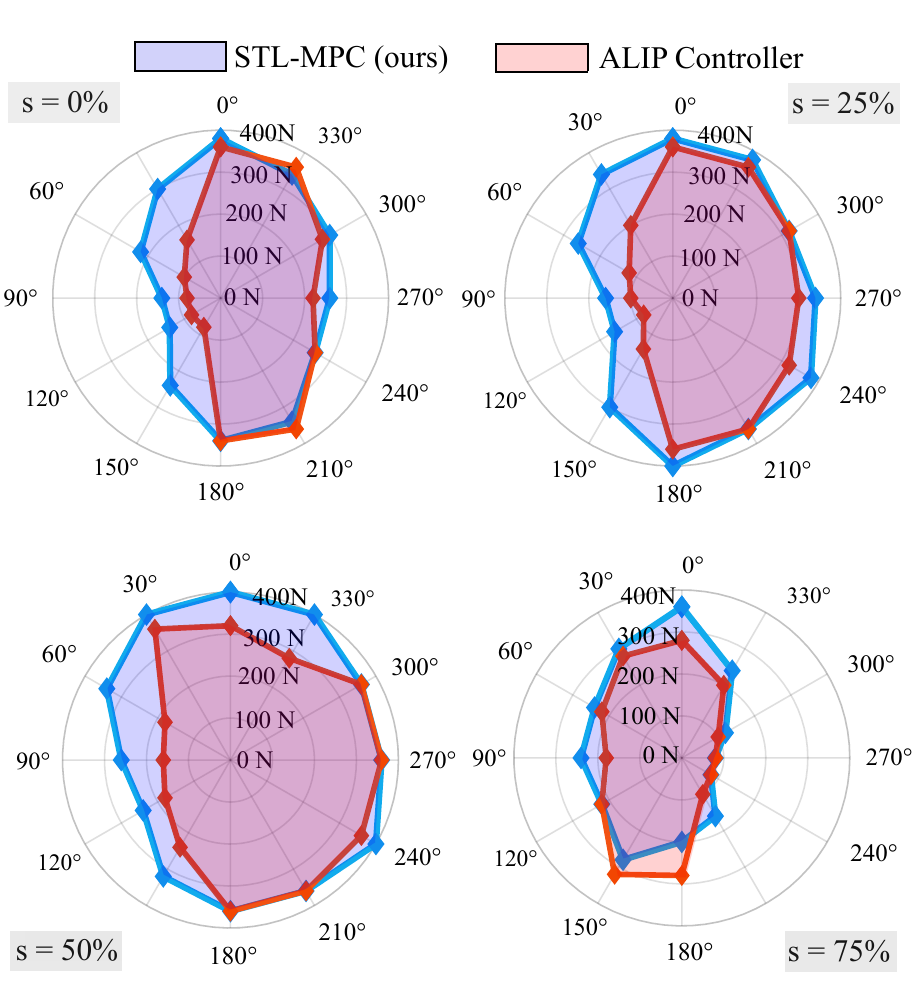}}
\caption{
The maximum force exerted on the pelvis from which the robot can safely recover within two steps in all 12 directions. The perturbations happen at different phases $s$ during a left leg stance. Values on the left half result in crossed-leg maneuvers, and values on the right half correspond to wide-step recoveries.}
\label{fig:spider}
\vspace{-0.25in}
\end{figure}

In Fig.~\ref{fig:spider}, we compare the maximum impulse the STL-MPC can withstand to that of the baseline ALIP controller. The STL-MPC demonstrates superior perturbation recovery performance across the vast majority of directions and phases, as reflected by the blue region encompassing the red region. The improvement is particularly evident for directions between $30^\circ$ and $150^\circ$, wherein crossed-leg maneuvers are induced for recovery, and active self-collision avoidance plays a critical role. This highlights the STL-MPC's capability to generate safe crossed-leg behaviors, thereby significantly enhancing its robustness against lateral perturbations. On the other hand, for perturbations between $210^\circ$ and $330^\circ$, both frameworks exhibit comparable performance, generating wide side-steps for recovery.
Note that we use $N=2$ walking steps as the MPC horizon, as existing studies \cite{two_step_enough, Koolen_capturability, Hierar_Opti_time} indicate that a two-step motion is sufficient for recovery to a periodic orbit.

Additionally, we observe the STL-MPC struggles most when the perturbation happens close to the end of a walking step at $s = 75\%$, as indicated by the smaller region than the baseline in the bottom right of Fig.~\ref{fig:spider}. This is due to the reduced flexibility to adjust the contact location and time within the short remaining duration of the perturbed step. 

\subsection{Stepping Stone Maneuvering}
\begin{figure}[t]
    \centering
    \includegraphics[width=0.45\textwidth]{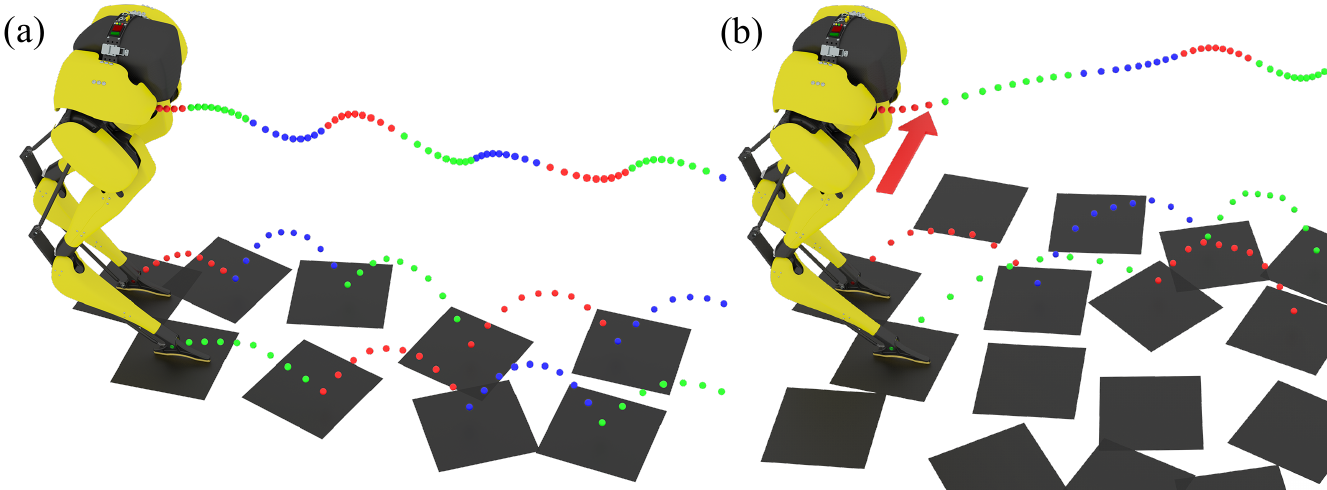}
    \caption{Illustration of maneuvering over two stepping-stone scenarios. (a) STL-MPC solves dynamically feasible trajectories that satisfy an additional foot-on-stones specification. (b) STL-MPC successfully plans crossed-leg maneuvers to recover from perturbation.}
    \label{fig:stepping_stone}
    \vspace{-0.20in}
\end{figure}

To demonstrate the STL-MPC's ability to handle a broad set of task specifications, we study locomotion in a stepping-stone scenario as shown in Fig.~\ref{fig:stepping_stone}. To restrict the foot location to the stepping stones, we augment the locomotion specification $\varphi_{\rm loco}$ with an additional specification $\varphi_{\rm stones}$ that encodes stepping stone locations. 
For each rectangular stone, the presence of a stance foot $\boldsymbol{p}_{\rm stance}$ inside its four edges is specified as 
$\varphi_{\rm stone}^s = \bigwedge_{i=1}^{4} (\mu^s_i(\boldsymbol{p}_{\rm stance}) \ge 0)$
, where $s \in \{1,\ldots,S\}$, $S$ is the total number of stepping stones, and $\mu^s_i$ is the signed distance from the stance foot to the $i^{\rm th}$ edge of the $s^{\rm th}$ stone. 
Then the combined foot location specification for $N$ walking steps is:
$$\varphi_{\rm stones} = \bigwedge_{j=1}^{N}(\square_{[T^j, T^j]}\bigvee_{s=1}^{S} \varphi_{\rm stone}^{s})$$
The augmented specification is the compound of the original locomotion specification $\varphi_{\rm loco}$ and the newly-added stepping stone specification:
$\varphi_{\rm loco}' = \varphi_{\rm loco} \wedge \varphi_{\rm stones}$.

We test STL-MPC using $\varphi_{\rm loco}'$ in two scenarios. The first scenario has stepping stones generated at ground level with random offsets and yaw rotations, as shown in Fig.~\ref{fig:stepping_stone}(a). 
The STL-MPC advances Cassie forward successfully.
In the second scenario, the STL-MPC demonstrates the ability to cross legs in response to a lateral perturbation in Fig.~\ref{fig:stepping_stone}(b). 


\subsection{Computation Speed Comparison between Smooth Encoding Method and Mixed-Integer Program}
To encode the robustness degree (as discussed in Sec.~\ref{Sec:STL}) of STL specifications into our gradient-based trajectory optimization (TO) formulation, we adopt a smooth-operator method \cite{smooth_operator} that allows a smooth gradient for efficient computation. Specifically, we replace the non-smooth ${\rm min}$ and ${\rm max}$ operators in the robustness degree (as defined in Table~II) with their smooth counterpart $\widetilde{{\rm min}}$ and $\widetilde{{\rm max}}$, as detailed in \cite{smooth_operator}. 

\begin{figure}[h]
\centering
\includegraphics[width=0.48\textwidth]{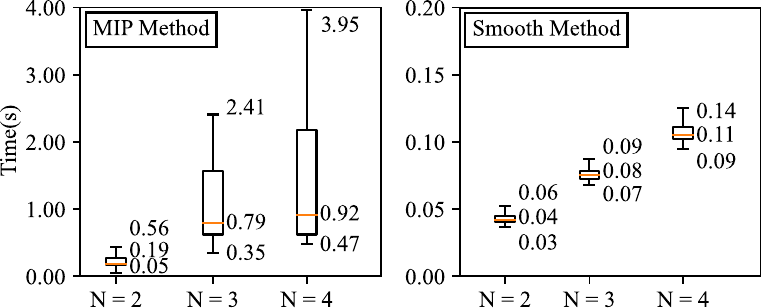}
\caption{
A comparison of the traditional MIP method and our smooth method shows the planning time to solve trajectories for $N$-walking-step horizons. The smooth method is faster and more consistent over all horizons.}
\label{fig:MIPvsSmooth}
\vspace{-0.15in}
\end{figure}

We benchmark the solving speed of the smooth method with the traditional mixed-integer programming (MIP) method \cite{STL_MPC_Raman}. The smooth method demonstrates a faster and more consistent solving speed, and its time consumption is nearer to linear with respect to the walking steps $N$. 

\section{Conclusion}
\label{Sec:conclusion}
This study presents a model predictive controller using signal temporal logic (STL) for bipedal locomotion push recovery. Our main contribution is the design of STL specifications that quantify the locomotion robustness and guarantee stable walking.
Our framework increased Cassie's impulse tolerance by $81$\% in critical crossed-leg scenarios. 
Further research will be focused on hardware verification and extensions to rough, dynamic terrain.

\clearpage 
\bibliographystyle{IEEEtran}
\bibliography{references}

\end{document}